\documentclass[runningheads]{llncs}
\usepackage{graphicx} 
\usepackage{amsmath,amssymb} 
\usepackage{color}

\usepackage[width=122mm,left=12mm,paperwidth=146mm,height=193mm,top=12mm,paperheight=217mm]{geometry}

\usepackage{algorithm} 
\usepackage{algpseudocode}
\usepackage{array}
\usepackage{multirow}

\usepackage[title]{appendix}
\newcommand{\tabincell}[2]{\begin{tabular}{@{}#1@{}}#2\end{tabular}}

\begin{document}
\pagestyle{headings}
\mainmatter

\title{Graph-based Heuristic Search \\for Module Selection Procedure \\in Neural Module Network}
\titlerunning{Graph-based Heuristic Search for Module Selection in NMN}
\author{Yuxuan Wu \and Hideki Nakayama}
\authorrunning{Y. Wu \and H. Nakayama}
\institute{The University of Tokyo\\
\email{\{wuyuxuan,nakayama\}@nlab.ci.i.u-tokyo.ac.jp}}

\maketitle

\begin{abstract}
Neural Module Network (NMN) is a machine learning model for solving the visual question answering tasks. NMN uses programs to encode modules' structures, and its modularized architecture enables it to solve logical problems more reasonably. However, because of the non-differentiable procedure of module selection, NMN is hard to be trained end-to-end. To overcome this problem, existing work either included ground-truth program into training data or applied reinforcement learning to explore the program. However, both of these methods still have weaknesses. In consideration of this, we proposed a new learning framework for NMN. Graph-based Heuristic Search is the algorithm we proposed to discover the optimal program through a heuristic search on the data structure named Program Graph. Our experiments on FigureQA and CLEVR dataset show that our methods can realize the training of NMN without ground-truth programs and achieve superior efficiency over existing reinforcement learning methods in program exploration.
\end{abstract}

\section{Introduction}

With the development of machine learning in recent years, more and more tasks have been accomplished such as image classification, object detection, and machine translation. 
However, there are still many tasks that human beings perform much better than machine learning systems, especially those in need of logical reasoning ability.
Neural Module Network (NMN) is a model proposed recently targeted to solve these reasoning tasks~\cite{nmn_0,nmn_1}.  
It first predicts a program indicating the required modules and their layout, and then constructs a complete network with these modules to accomplish the reasoning. 
With the ability to break down complicated tasks into basic logical units and to reuse previous knowledge, NMN achieved super-human level performance on challenging visual reasoning tasks like CLEVR~\cite{clevr_iep}. 
However, because the module selection is a discrete and non-differentiable process, it is not easy to train NMN end-to-end. 

To deal with this problem, 
a general solution is to separate the training into two parts: the program predictor and the modules. 
In this case, the program becomes a necessary intermediate label. The two common solutions to provide this program label are either to include the ground-truth programs into training data or to apply reinforcement learning to explore the optimal candidate program.
However, these two solutions still have the following limitations. 
The dependency on ground-truth program annotation makes NMN's application hard to be extended to datasets without this kind of annotation. 
This annotation is also highly expensive while being hand-made by humans. Therefore, program annotation cannot always be expected to be available for tasks in real-world environments. 
In view of this, methods relying on ground-truth program annotation cannot be considered as complete solutions for training NMN. On the other hand, the main problem in the approaches based on reinforcement learning is that with the growth of the length of programs and number of modules,
the size of the search space of possible programs becomes so huge that a reasonable program may not be found in an acceptable time. 

In consideration of this, we still regard the training of NMN as an open problem. 
With the motivation to take advantage of NMN on broader tasks and overcome the difficulty in its training in the meanwhile, in this work, we proposed a new learning framework to solve the non-differentiable module selection problem in NMN. 

\begin{figure}[tb]
\centering
\includegraphics[width=\textwidth]{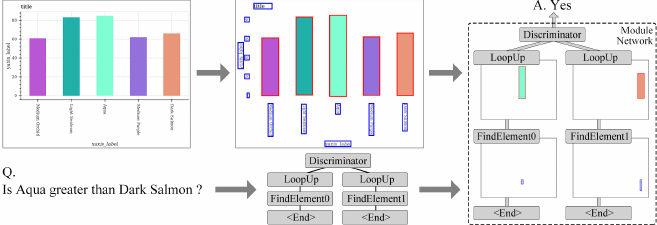}
\caption{Our learning framework enables the NMN to solve the visual reasoning problem without ground-truth program annotation.}
\label{fig:introduction}
\end{figure}

In this learning framework, 
we put forward the Graph-based Heuristic Search algorithm to enable the model to find the most appropriate program by itself. Basically, this algorithm is inspired by Monte Carlo Tree Search (MCTS). Similar to MCTS, our algorithm conducts a heuristic search to discover the most appropriate program in the space of possible programs. Besides, inspired by the intrinsic connection between programs, we proposed the data structure named Program Graph to represent the space of possible programs in a way more reasonable than the tree structure used by MCTS. 
Further, to deal with the cases that the search space is extremely huge, we proposed the Candidate Selection Mechanism to narrow down the search space.

With these proposed methods, our learning framework implemented the training of NMN regardless of the existence of the non-differentiable module selection procedure.
Compared to existing work, our proposed learning framework has the following notable characteristics:

\begin{itemize}
  \item It can implement the training of NMN with only the triplets of \{question, image, answer\} and without the ground-truth program annotation.
  \item It can explore larger search spaces more reasonably and efficiently.
  \item It can work on both trainable modules with neural architectures and non-trainable modules with discrete processing.
\end{itemize}


\section{Related Work}

\subsection{Visual Reasoning}\label{subsec:visual_reasoning}

Generally, Visual Reasoning can be considered as a kind of Visual Question Answering (VQA)~\cite{vqa}. 
Besides the requirement of understanding information from both images and questions in common VQA problems, 
Visual Reasoning further asks for the capacity to recognize abstract concepts such as spatial, mathematical, and logical relationships. 
CLEVR~\cite{clevr} is one of the most famous and widely used datasets for Visual Reasoning. It provides not only the triplets of \{question, image, answer\} but also the functional programs paired with each question.
FigureQA~\cite{figureqa} is another Visual Reasoning dataset we focus on in this work. It provides questions in fifteen different templates asked on five different types of figures.

To solve Visual Reasoning problems, a naive approach would be the combination of Convolutional Neural Network (CNN) and Recurrent Neural Network (RNN). Here, CNN and RNN are responsible for extracting information from images and questions, respectively. Then, the extracted information is combined and fed to a decoder to obtain the final answer. However, this methodology of treating Visual Reasoning simply as a classification problem sometimes cannot achieve desirable performance due to the difficulty of learning abstract concepts and relations between objects~\cite{vqa,figureqa,clevr_iep}. 
Instead, more recent work applied models based on NMN to solve Visual Reasoning problems~\cite{clevr_iep,att_nmn_0,att_nmn_1,att_nmn_2,obj_nmn_0,obj_nmn_1,nscl}.

\subsection{Neural Module Network}\label{subsec:nmn}

Neural Module Network (NMN) is a machine learning model proposed in 2016~\cite{nmn_0,nmn_1}. 
Generally, the overall architecture of NMN can be considered as a controller and a set of modules. 
Given the question and the image, firstly, the controller of NMN takes the question as input and outputs a program indicating the required modules and their layout. Then, the specified modules are concatenated with each other to construct a complete network. Finally, the image is fed to the assembled network and the answer is acquired from the root module. 
As far as we are concerned, the advantage of NMN can be attributed to the ability to break down complicated questions into basic logical units and the ability to reuse previous knowledge efficiently.

By the architecture of modules, NMN can further be categorized into three subclasses: the feature-based, attention-based, and object-based NMN. 

For feature-based NMNs, the modules apply CNNs and their calculations are directly conducted on the feature maps. 
Feature-based NMNs are the most concise implementation of NMN and were utilized most in early work~\cite{clevr_iep}.

For attention-based NMNs, the modules also apply neural networks but their calculations are conducted on the attention maps. 
Compared to feature-based NMNs, attention-based NMNs retain the original information within images better so they achieved higher reasoning precision and accuracy~\cite{nmn_0,nmn_1,att_nmn_0,att_nmn_2}.

For object-based NMNs, they regard the information in an image as a set of discrete representations on objects instead of a continuous feature map. Correspondingly, their modules conduct pre-defined discrete calculations. Compared to feature-based and attention-based NMNs, object-based NMNs achieved the highest precision on reasoning~\cite{obj_nmn_0,obj_nmn_1}. However, their discrete design usually requires more prior knowledge and pre-defined attributes on objects.

\subsection{Monte Carlo Methods}\label{subsec:mcm}

Monte Carlo Method is the general name of a group of algorithms that make use of random sampling to get an approximate estimation for a numerical computing~\cite{mcm}. 
These methods are broadly applied to the tasks that are impossible or too time-consuming to get exact results through deterministic algorithms. 
Monte Carlo Tree Search (MCTS) is an algorithm that applied the Monte Carlo Method to the decision making in game playing like computer Go~\cite{mcts_0,mcts_1}. Generally, this algorithm arranges the possible state space of games into tree structures, and then applies Monte Carlo estimation to determine the action to take at each round of games.
In recent years, there also appeared approaches to establish collaborations between Deep Learning and MCTS. These work, represented by AlphaGo, have beaten top-level human players on Go, which is considered to be one of the most challenging games for computer programs~\cite{alphago_0,alphago_1}.


\section{Proposed Method}\label{sec:method}

\subsection{Overall Architecture}\label{subsec:method_overall}
\begin{figure}[tb]
\centering
\includegraphics[width=\textwidth]{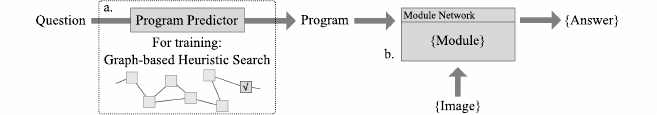}
\caption{Our Graph-based Heuristic Search algorithm assists the learning of the Program Predictor.}
\label{fig:method_general}
\end{figure}

The general architecture of our learning framework is shown as Fig.\ref{fig:method_general}.
As stated above, the training of the whole model can be divided into two parts: a. Program Predictor and b. modules. The main difficulty of training comes from the side of Program Predictor because of the lack of 
expected programs as training labels.
To overcome this difficulty, we proposed the algorithm named Graph-based Heuristic Search to enable the model to find the optimal program by itself through a heuristic search on the data structure Program Graph. After this searching process, the most appropriate program that was found is utilized as the program label so that the Program Predictor can be trained in a supervised manner. In other words, this searching process can be considered as a procedure targeted to provide training labels for the Program Predictor. 

The abstract of the total training workflow is presented as Algorithm \ref{alg:total}. Note that here 
$q$ denotes the question, $p$ denotes the program, \{$module$\} denotes the set of modules available in the current task, 
\{$img$\} denotes the set of images that the question is asking on, \{$ans$\} denotes the set of answers paired with images.
Details about the $Sample$ function are provided in Appendix A.

\begin{algorithm}
\caption{Total Training Workflow}
\begin{algorithmic}[1]
\Function{Train()}{}
    \State Program\_Predictor, \{$module$\} $\gets$ Intialize()
    \For{loop in range($Max\_loop$)}
        \State $q$, \{$img$\}, \{$ans$\} $\gets$ Sample(Dataset)
        \State $p$ $\gets$ Graph-based\_Heuristic\_Search($q$, \{$img$\}, \{$ans$\}, \{$module$\})
        \State Program\_Predictor.train($q$, $p$)
    \EndFor
\EndFunction
\end{algorithmic}
\label{alg:total}
\end{algorithm}

\subsection{Program Graph}\label{subsec:program_graph}

To start with, we first give a precise definition of the program we use.
Note that each of the available modules in the model has a unique name, fixed numbers of inputs, and one output. Therefore, a program can be defined as a tree meeting the following rules :

i) Each of the non-leaf nodes stands for a possible module, each of the leaf nodes holds a $\langle$END$\rangle$ flag.

ii) The number of children that a node has equal to the number of inputs of the module that the node represents.

For the convenience of representation in prediction, a program can also be transformed into a sequence of modules together with $\langle$END$\rangle$ flags via pre-order tree traversal. Considering that the number of inputs of each module is fixed, the tree form can be rebuilt from such sequence uniquely.

Then, as for the Program Graph, 
Program Graph is the data structure we use to represent the relation between all programs that have been reached throughout the searching process, and it is also the data structure that our algorithm Graph-based Heuristic Search works on. A Program Graph can be built meeting the following rules :

i) Each graph node represents a unique program that has been reached.

ii) There is an edge between two nodes if and only if the edit distance of their programs is one. Here, insertion, deletion, and substitution are the three basic edit operations whose edit distance is defined as one. 
Note that the edit distance between programs is judged on their tree form. 

iii) Each node in the graph maintains a score. 
This score is initialized as the output probability of the program of a node according to the Program Predictor when the node is created,
and can be updated when the program of a node is executed.

\begin{figure}[tb]
\centering
\includegraphics[width=\textwidth]{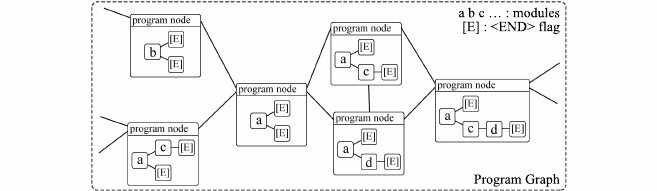}
\caption{Illustration of part of a Program Graph}
\label{fig:program_graph}
\end{figure}

Fig.\ref{fig:program_graph} is an illustration of a Program Graph consisting of several program nodes together with their program trees as examples. 
To distinguish the node in the tree of a program and the node in the Program Graph, the former will be referred to as $m\_n$ for ``module node" and the latter will be referred to as $p\_n$ for ``program node" in the following discussion. 
Details about the initialization of the Program Graph are provided in Appendix B.

\subsection{Graph-based Heuristic Search}\label{subsec:gbhs}
 
\begin{algorithm}[htb]
\caption{Graph-based Heuristic Search}
\begin{algorithmic}[1]
\Function{Main($q$, \{$img$\}, \{$ans$\}, \{$module$\})}{}
    \State $g$ $\gets$ InitializeGraph($q$) 
    \For{step in range($Max\_step$)}
        \State \{$p\_n$\}$_c$ $\gets$ $p\_n$ \textbf{for} $p\_n$ in $g$ \textbf{and} $p\_n$.fully\_explored == False
        \State $p\_n_i$.Exp $\gets$ FindExpectation($p\_n_i$, $g$) \textbf{for} $p\_n_i$ in \{$p\_n$\}$_c$
        \State $p\_n_e$ $\gets$ $p\_n_i$ s.t. $p\_n_i$.Exp = $max$\{$p\_n_i$.Exp \textbf{for} $p\_n_i$ in \{$p\_n$\}$_c$\} 
        \State Expand($p\_n_e$, $g$, \{$img$\}, \{$ans$\}, \{$module$\})
    \EndFor
    \State $p\_n_{best}$ $\gets$ $p\_n_i$ s.t. $p\_n_i$.score = $max$\{$p\_n_i$.score \textbf{for} $p\_n_i$ in \{$p\_n_i$\}\} 
    \State \Return $p\_n_{best}$.program
\EndFunction
\item[]
\Function{Expand}{$p\_n_e$, $g$, \{$img$\}, \{$ans$\}, \{$module$\}}
    \State $p\_n_e$.\rm{visit\_count} $\gets$ $p\_n_e$.\rm{visit\_count} + 1
    \If{$p\_n_e$.visited == False}
        \State $p\_n_e$.score $\gets$ accuracy($p\_n_e$.program, \{$img$\}, \{$ans$\}, \{$module$\})
        \State $p\_n_e$.visited $\gets$ True
    \EndIf
    \State \{$m\_n$\}$_c$ $\gets$ $m\_n$ \textbf{for} $m\_n$ in $p\_n_e$.program \textbf{and} $m\_n$.expanded == False
    \State $m\_n_m$ $\gets$ Sample(\{$m\_n$\}$_c$)
    \State \{$\rm{program}$\}$_{new}$ $\gets$ Mutate($p\_n_e$.program, $m\_n_m$, \{$module$\})
    \For{$\rm{program}$$_i$ in \{$\rm{program}$\}$_{new}$}
        \If{LegalityCheck($\rm{program}$$_i$) == True}
            \State $g$.update($\rm{program}$$_i$)
        \EndIf
    \EndFor
    \State $m\_n_m$.expanded $\gets$ True
    \State $p\_n_e$.fully\_explored $\gets$ True \textbf{if} \{$m\_n$\}$_c$.remove($m\_n_m$) == $\varnothing$
\EndFunction
\end{algorithmic}
\label{alg:gbhs}
\end{algorithm}

Graph-based Heuristic Search is the core algorithm in our proposed learning framework. Its basic workflow is presented as the $Main$ function in line 1 of Algorithm \ref{alg:gbhs}. After Program Graph $g$ gets initialized, the basic workflow can be described as a recurrent exploration on the Program Graph consisting of the following four steps : 

i) Collecting all the program nodes in Program Graph $g$ that have not been fully explored yet as the set of candidate nodes \{$p\_n$\}$_c$.

ii) Calculating the Expectation for all the candidate nodes.

iii) Selecting the node with the highest Expectation value among all the candidate nodes.

iv) Expanding on the selected node to generate new program nodes and update the Program Graph. 

The details about the calculation of Expectation and expanding strategy are as follows.

\subsubsection{Expectation}\label{subsub:exp_score}

Expectation is a grade defined on each program node to determine which node should be selected for the following expansion step. 
This Expectation is calculated through the following Equation \ref{eq:exp}.

\begin{equation}
\begin{split}
\mathrm{Exp} = & \sum\limits_{d = 0}^{D} w_d * max\{p\_n_j. {\mathrm{score}} \mid p\_n_j\ in\ g,\ distance(p\_n_i, p\_n_j) \leq d\} \\
 & + \frac{\alpha}{p\_n_i.{\mathrm{visit\_count}} + 1}
\label{eq:exp}
\end{split}
\end{equation}

Intuitively, this equation measures how desirable a program is to guide the modules to answer a given question reasonably. 
Here, $D$, $w_d$, and $\alpha$ are hyperparameters indicating the max distance in consideration, a sequence of weight coefficients while summing best scores in the different distance $d$, and the scale coefficient to encourage visiting unexplored nodes, respectively. 

In this equation, the first term observes the nodes nearby and find the highest score in each different distance $d$ from $0$ to $D$. Then, these scores are weighted by $w_d$ and summed up. Note that the distance here is measured on the Program Graph, which also equals to the edit distance between two programs. 
The second term in this equation is a balance term negatively correlated to the number of times that a node has been visited and expanded on. 
This term balances the grades of unexplored or less explored nodes. 

\subsubsection{Expansion Strategy}\label{subsub:exp_stg}

Expansion is another important procedure in our proposed algorithm as shown in line 12 of Algorithm \ref{alg:gbhs}.
The main objective of this procedure is to generate new program nodes and update the Program Graph. To realize this, the five main steps are as follows:

i) If the node $p\_n_e$ in Program Graph is visited for the first time, try its program by building the model with specified modules to answer the question, then update the score of the node with the accuracy. If there are modules with neural architecture, these modules should also be trained here, but the updated parameters are retained only if the new accuracy exceeds the previous one.

ii) Collect the module nodes that have not been expanded on yet within the program, then sample one from them as the module node $m\_n_m$ to expand on. 

iii) Mutate the program at module $m\_n_m$ to generate a new set of programs \{$\text{program}$\}$_{new}$ with three edit operations: insertion, deletion, and substitution.

iv) For the new programs judged to be legal, 
if there is not yet a node representing the same program in the Program Graph $g$, then create a new program node representing this program and add it to $g$. The related edge should also be added to $g$ if it does not exist yet.

v) If all of the module nodes have been expanded on, then mark this program node $p\_n_e$ as fully explored.

\begin{figure}[htb]
\centering
\includegraphics[width=\textwidth]{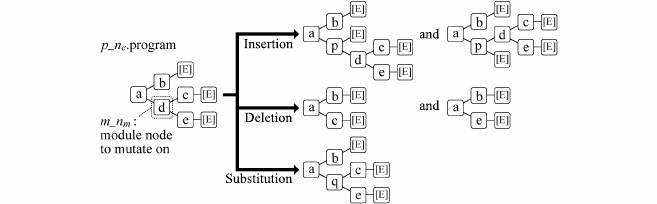}
\caption{Example of the mutations generated by the three opeartions insertion, deletion, and subsitution.}
\label{fig:mutate}
\end{figure}

For the Mutation in step iii), the three edit operations are illustrated by Fig.\ref{fig:mutate}. 
Here, insertion adds a new module node between the node $m\_n_m$ and its parent node. The new module can be any of the available modules in the model. If the new module has more than one inputs, $m\_n_m$ should be set as one of its children, and the rest of the children are set to leaf nodes with $\langle$END$\rangle$ flag. 

Deletion deletes the node $m\_n_m$ and set its child as the new child of $m\_n_m$'s parent. If $m\_n_m$ has more than one child, only one of them should be retained and the others are abandoned. 

Substitution replaces the module of $m\_n_m$ with another module. The new module can be any of the modules that have the same number of inputs as $m\_n_m$. 

For insertion and deletion, if they are multiple possible mutations because the related node has more than one child as shown in Fig.\ref{fig:mutate}, all of them are retained. 

These rules ensure that newly generated programs consequentially have legal structures, but there are still cases that these programs are not legal in the sense of semantics, e.g., the output data type of a module does not match the input data type of its parent. 
Legality check is conducted to determine whether a program is legal and should be added to the Program Graph, more details about this function are provided in Appendix C.

\subsection{Candidate Selection Mechanism for Modules}\label{subsec:method_candidate}
The learning framework presented above is already a complete framework to realize the training of the NMN. However, in practice we found that with the growth of the length of programs and the number of modules, the size of search space explodes exponentially. This brings trouble to the search. To overcome this problem, 
we further proposed the Candidate Selection Mechanism (CSM), which is an optional component within our learning framework. 
Generally speaking, if CSM is activated, it selects only a subset of modules from the whole of available modules. Then, only these selected modules are used in the following Graph-based Heuristic Search. 
The abstract of the training workflow with CSM is presented as Algorithm \ref{alg:with_cand}.

\begin{algorithm}[b]
\caption{Training Workflow with Candidate Selection Mechanism}
\begin{algorithmic}[1]
\Function{Train()}{}
    \State Program\_Predictor, Necessity\_Predictor, \{$module$\} $\gets$ Intialize()
    \For{loop in range($Max\_loop$)}
        \State $q$, \{$img$\}, \{$ans$\} $\gets$ Sample(Dataset)
        \State \{$module$\}$_{candidate}$ $\gets$ Necessity\_Predictor($q$, \{$module$\})
        \State $p$ $\gets$ Graph-based\_Heuristic\_Search($q$, \{$img$\}, \{$ans$\}, \{$module$\}$_{candidate}$)
        \State Necessity\_Predictor.train($q$, $p$)
        \State Program\_Predictor.train($q$, $p$)
    \EndFor
\EndFunction
\end{algorithmic}
\label{alg:with_cand}
\end{algorithm}

\begin{figure}[h]
\centering
\includegraphics[width=\textwidth]{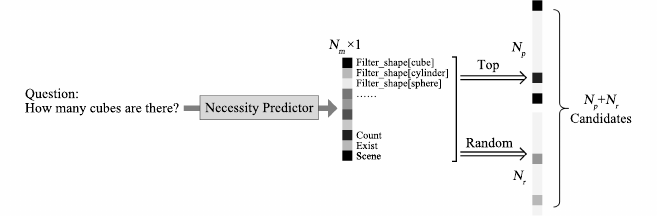}
\caption{The process to selecte the $N_p + N_r$ candidate modules}
\label{fig:candiate}
\end{figure}

Here, we included another model named Necessity Predictor into the learning framework. This model takes the question as input, and predicts a $N_m$-dimensions vector as shown in Fig.\ref{fig:candiate}. Here, $N_m$ indicates the total number of modules. Each value in the output vector is a real number in the range of [0, 1] indicating the possibility that each module is necessary for the solution of the given question. 
$N_p$ and $N_r$ are the two hyperparameters for the candidate modules selection procedure. 
$N_p$ indicates the number of modules that are selected according to the predicted possibility value, i.e., 
to select $N_p$ modules with the top $N_p$ prediction values.
$N_r$ indicates the number of modules that are selected randomly besides the $N_p$ ones. 
Then, the union of these two selections with $N_p + N_r$ modules becomes the candidate modules for the following search.

For the training of this Necessity Predictor, the best program found in the search is transformed into a $N_m$-dimensions boolean vector indicating whether each module appeared in the program. Then, this boolean vector is set as the training label so that the Necessity Predictor can also be trained in a supervised manner as Program Predictor does.

\section{Experiments and Results}\label{sec:experiment}

Our experiments are conducted on the FigureQA and the CLEVR dataset. Their settings and results are presented in the following subsections respectively. 

\subsection{FigureQA Dataset}
The main purpose of the experiment on FigureQA is to certify that our learning framework can realize the training of NMN on a dataset without ground-truth program annotations and outperform the existing methods with models other than NMN. 

\begin{figure}[b]
\centering
\includegraphics[width=\textwidth]{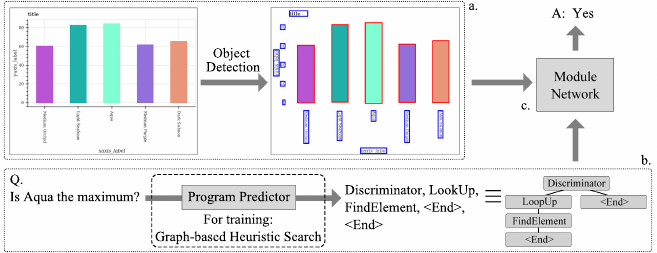}
\caption{An example of the inference process on FigureQA}
\label{fig:figureqa_overview}
\end{figure}

An overview of how our methods work on this dataset is shown in Fig.\ref{fig:figureqa_overview}. 
Considering that the size of the search space of the programs used in FigureQA is relatively small, the CSM introduced in Section \ref{subsec:method_candidate} is not activated.

Generally, the workflow consists of three main parts. 
Firstly, the technique of 
object detection~\cite{rcnn} 
together with 
optical character recognition~\cite{ocr_survey} 
are applied to transform the raw image into discrete element representations as shown in Fig.\ref{fig:figureqa_overview}.a. 
For this part, we applied Faster R-CNN~\cite{faster_rcnn_0,faster_rcnn_1} with ResNet 101 as the backbone for object detection and Tesseract open source OCR engine~\cite{tesseract,github_tesseract} for text recognition. All the images are resized to 256 by 256 pixels before following calculations.

Secondly, for the part of program prediction as shown in Fig.\ref{fig:figureqa_overview}.b., we applied our Graph-based Heuristic Search algorithm for the training. 
The setting of the hyperparameters for this part are shown in Table \ref{tab:figureqa_hyper}.
The type of figure is treated as an additional token appended to the question.

Thirdly, for the part of modules as shown in Fig.\ref{fig:figureqa_overview}.c., 
we designed some pre-defined modules with discrete calculations on objects. Their functions are corresponded to the reasoning abilities required by FigureQA. 
These pre-defined modules are used associatively with modules with neural architecture. 
Details of all these modules are provided in Appendix D.

Table \ref{tab:result_total} shows the results of our methods compared with baseline and existing methods. ``Ours" is the primitive result from the experiment settings presented above. 
Besides, we also provide the result named ``Ours + GE" where ``GE" stands for ground-truth elements. In this case, element annotations are obtained directly from ground-truth plotting annotations provided by FigureQA instead of the object detection results. We applied this experiment setting to measure the influence of the noise in object detection results.

\begin{table}[t]
\caption{Setting of hyperparameters in our experiment}
\centering
\begin{tabular}{p{50pt}<{\centering} p{50pt}<{\centering} p{20pt}<{\centering} p{100pt}<{\centering} p{25pt}<{\centering}}
\hline
$Max\_loop$ & $Max\_step$ & $D$ & $w_d$ & $\alpha$ \\
\hline
100 & 1000 & 4 & (0.5, 0.25, 0.15, 0.1) & 0.05 \\
\hline
\end{tabular}
\label{tab:figureqa_hyper}
\end{table}

\begin{table}[t]
\caption{Comparison of accuracy with previous methods on the FigureQA dataset.}
\centering
\begin{tabular}{p{108pt}<{\centering} p{40pt}<{\centering} p{40pt}<{\centering} c p{40pt}<{\centering} p{40pt}<{\centering} }\hline
\multirow{3}{*}{Method} & \multicolumn{5}{c}{Accuracy} \\
\cline{2-6}
& \multicolumn{2}{c}{Validation Sets} && \multicolumn{2}{c}{Test Sets} \\
\cline{2-3} \cline{5-6}
& Set 1 & Set 2 && Set 1 & Set 2 \\
\hline
Text only~\cite{figureqa} & \multicolumn{2}{c}{50.01\%} && \multicolumn{2}{c}{50.01\%} \\
CNN+LSTM~\cite{figureqa} & \multicolumn{2}{c}{56.16\%} && \multicolumn{2}{c}{56.00\%} \\
Relation Network~\cite{figureqa,relation_net} & \multicolumn{2}{c}{72.54\%} && \multicolumn{2}{c}{72.40\%} \\
Human~\cite{figureqa} & & && \multicolumn{2}{c}{91.21\%} \\
FigureNet~\cite{figurenet} & & && \multicolumn{2}{c}{84.29\%} \\
PTGRN~\cite{PTGRN} & \multicolumn{2}{c}{86.25\%} && \multicolumn{2}{c}{86.23\%} \\
PReFIL~\cite{PReFIL} & 94.84\% & 93.26\% && 94.88\% & 93.16\% \\
\hline
Ours & 95.74\% & 95.55\% && \textbf{95.61}\% & \textbf{95.28}\%  \\
Ours + GE & \textbf{96.61}\% & \textbf{96.52}\% && & \\
\hline
\end{tabular}
\label{tab:result_total}
\end{table}

Through the result, firstly it can be noticed that both our method and our method with GE outperform all the existing methods. In our consideration, the superiority of our method mainly comes from the successful application of NMN. As stated in Section \ref{subsec:nmn}, NMN has shown outstanding capacity in solving logical problems. However, limited by the non-differentiable module selection procedure, the application of NMN can hardly be extended to those tasks without ground-truth program annotations like FigureQA. In our work, the learning framework we proposed can realize the training of NMN without ground-truth programs so that we succeeded to apply NMN on this FigureQA. This observation can also be certified through the comparison between our results and PReFIL. 

Compared to PReFIL, considering that we applied the nearly same 40-layer DenseNet to process the image, the main difference we made in our model is the application of modules. The modules besides the final Discriminator ensure that the inputs fed to the Discriminator are related to what the question is asking on more closely.

Here, another interesting fact shown by the result is the difference between accuracies reached on set 1 and set 2 of both validation sets and test sets. Note that in FigureQA, validation set 1 and test set 1 adopted the same color scheme as the training set, while validation set 2 and test set 2 adopted an alternated color scheme. This difference leads to the difficulty of the generalization from the training set to the two set 2. As a result, for PReFIL the accuracy on each set 2 drops more than 1.5\% from the corresponding set 1. However, for our method with NMN, this decrease is only less than 0.4\%, which shows a better generalization capacity brought by the successful application of NMN.

Also, Appendix E reports the accuracies achieved on test set 2 by different question types and figure types. It is worth mentioning that our work is the first one to exceed human performance on every question type and figure type.

\subsection{CLEVR Dataset}

The main purpose of the experiment on CLEVR is to certify that 
our learning framework can achieve superior searching efficiency compared to the classic reinforcement learning method. 

For this experiment, we created a subset of CLEVR containing only those training data whose questions appear at least two times in all training questions. There are 31252 different questions together with their corresponding programs in this subset. 
The reason of applying such a subset is that the size of the whole space of possible programs is approximately up to $10^{40}$, which is so huge that no existing method can realize the search in it without any prior knowledge or simplification on programs.
Considering that the training of modules is highly time-consuming, we only activate the part of program prediction in our learning framework, which is shown as Fig.\ref{fig:figureqa_overview}.b. With this setting, the modules specified by the program would not be trained actually. Instead, a boolean value indicating whether the program is correct or not is returned to the model as a substitute for the question answering accuracy. Here, only the programs that are exactly the same as the ground-truth programs paired with given questions are considered as correct.

In this experiment, comparative experiments were made on the cases of both activating and not activating the CSM.
The structures of the models used as the Program Predictor and the Necessity Predictor are as follows. 
For Program Predictor, we applied a 2-layer Bidirectional LSTM with hidden state size of 256 as the encoder, and a 2-layer LSTM with hidden state size of 512 as the decoder. Both the input embedding size of encoder and decoder are 300. The setting of hyperparameters are the same as FigureQA as shown in Table.\ref{tab:figureqa_hyper} except that $Max\_loop$ is not limited.
For Necessity Predictor, we applied a 4-layer MLP. The input of the MLP is a boolean vector indicating whether each word in the dictionary appears in the question, the output of the MLP is a 39-dimensional vector for there are 39 modules in CLEVR, the size of all hidden layers is 256. The hyperparameters $N_p$ and $N_r$ are set to 15 and 5 respectively. For the sentence embedding model utilized in the initialization of the Program Graph, we applied the GenSen model with pre-trained weights~\cite{gensen,github_gensen}.

For the baseline, we applied REINFORCE~\cite{reinforce} as most of the existing work~\cite{clevr_iep,nscl} did to train the same Program Predictor model. 

\begin{figure}[t]
\centering
\includegraphics[width=\textwidth]{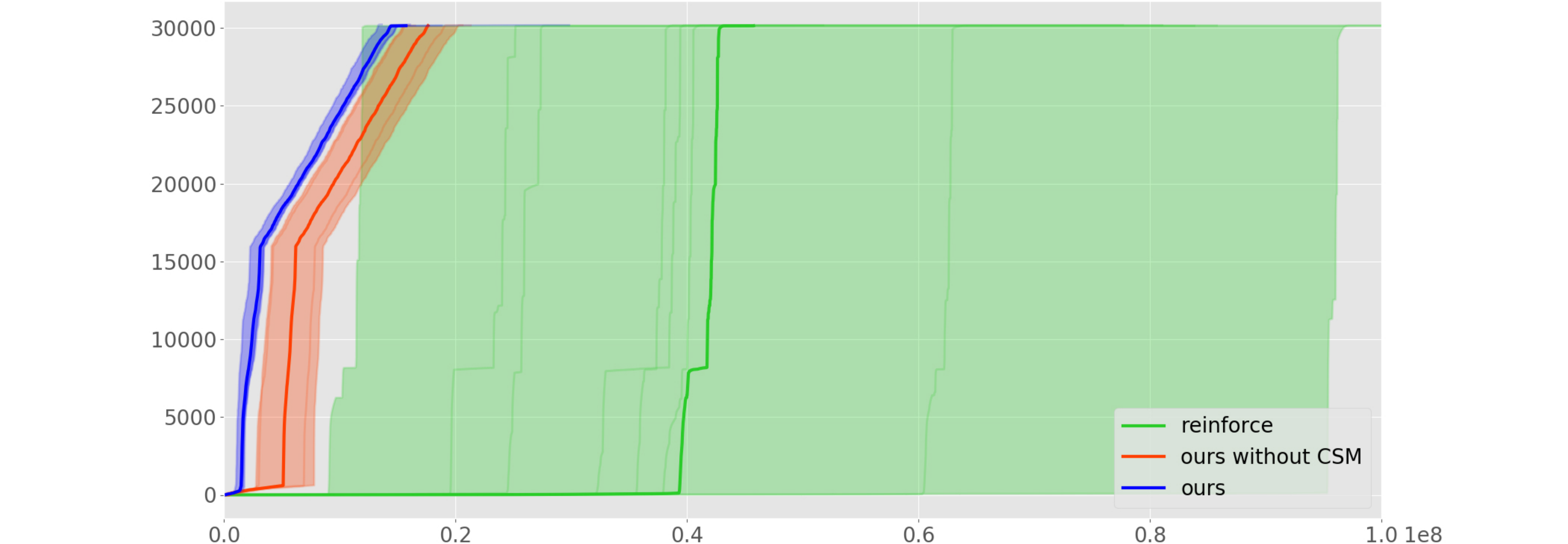}
\caption{Relation between the times of search and the number of correct programs found within the searching processes of three methods.}
\label{fig:clevr_result}
\end{figure}

The searching processes of our method, our method without CSM, and REINFORCE are shown by Fig.\ref{fig:clevr_result}. 
Note that in this figure, the horizontal axis indicates the times of search, the vertical axis indicates the number of correct programs found. 
The experiments on our method and our method without CSM are repeated four times each, and the experiment on REINFORCE is repeated eight times. 
Also, we show the average results as the thick solid lines in this figure indicating the average times of search used to find specific numbers of correct programs. 
Although in this subset of CLEVR, the numbers of correct programs that can be finally found are quite similar for the three methods, their searching processes show great differences.
From this result, three main conclusions can be drawn. 

Firstly, in terms of the average case, our method shows a significantly higher efficiency in searching appropriate programs.

Secondly, the searching process of our method is much more stable while the best case and worst case of REINFORCE differ greatly.

Thirdly, the comparison between the result of our method and our method without CSM certified the effectiveness of the CSM.

\section{Conclusion}\label{sec:conclusion}

In this work, to overcome the difficulty of training NMN because of its non-differentiable module selection procedure, we proposed a new learning framework for the training of the NMN. Our main contribution in this framework can be summarized as follows.

Firstly, we proposed the data structure named Program Graph to represent the search space of programs more reasonably. 

Secondly and most importantly, we proposed the Graph-based Heuristic Search algorithm to enable the model to find the most appropriate program by itself
to get rid of the dependency on the ground-truth programs in training.

Thirdly, we proposed the Candidate Selection Mechanism to improve the performance of the learning framework when the search space is huge.

Through the experiment, the experiment on FigureQA certified that our learning framework can realize the training of NMN on a dataset without ground-truth program annotations and outperform the existing methods with models other than NMN. The experiment on CLEVR certified that 
our learning framework can achieve superior efficiency in searching programs compared to the classic reinforcement learning method. 
In view of this evidence, we conclude that our proposed learning framework is a valid and advanced approach to realize the training of NMN.

Nevertheless, our learning framework still cannot deal with the extremely huge search spaces, e.g., the whole space of possible programs in CLEVR. We leave further study on methods that can realize the search in such enormous search spaces as the future work.

\section*{Acknowledgment}

This work was supported by JSPS KAKENHI Grant Number JP19K22861.


\bibliographystyle{splncs}
\bibliography{egbib}


\clearpage

\appendix
\begin{subappendices}
\renewcommand{\thesection}{\Alph{section}}

\section{Training Data Sampling}\label{apx:data_smp}

\begin{figure}[htb]
\centering
\includegraphics[width=\textwidth]{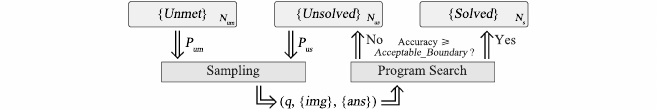}
\caption{Data sampling strategy}
\label{fig:data_sampling}
\end{figure}

The basic sampling unit of training data is triplet as ($q$, \{$img$\}, \{$ans$\}). 
Generally, as shown in Fig.\ref{fig:data_sampling}, we maintain three sets \{$Unmet$\}, \{$Unsolved$\}, \{$Solved$\} to distinguish training data in different status. 

Intuitively, \{$Unmet$\} contains training data that have not been met and used. At the beginning of learning, all the training data is stored in \{$Unmet$\}. 

\{$Unsolved$\} contains training data that has been sampled from \{$Unmet$\} but on which the final accuracy achieved in the following search did not reach a hyperparameter named $Acceptable\_Boundary$. 

\{$Solved$\} contains training data that has been sampled from \{$Unmet$\} or \{$Unsolved$\} and the final accuracy reached the $Acceptable\_Boundary$. 

We denote the numbers of training data triplets in these three sets as $N_{um}$, $N_{us}$ and $N_{s}$, respectively.

For each $Sample$ step in each training loop, the training data can be sampled from either \{$Unmet$\} or \{$Unsolved$\} with probability $P_{um}$ and $P_{us}$ as shown in Equation \ref{eq:samp}.

\begin{subequations}
\begin{align}
P_{um} = & \begin{cases}
    e^{- \frac{N_{us}}{N_{s} + 1}}, & \mathrm{if}\ N_{um} > 0\ ; \\
    0, & \mathrm{otherwise}
\end{cases}\\
P_{us} = & \ 1 - P_{um}
\end{align}
\label{eq:samp}
\end{subequations}

\section{Program Graph Initialization}\label{apx:graph_ini}

To initialize the Program Graph, at most three initial program nodes are created as the starting points for the following search. The programs of them are:

i) The program predicted by the Program Predictor model.

ii) The program found for the question within \{$Solved$\} that is closest to the current given question.

iii) The shortest legal program.

Specifically for ii), this term only works when \{$Solved$\} is not empty. If so, a pre-trained sentence embedding model $SE(\cdot)$ is utilized to judge the semantic distance between questions and find a question $q_{c}$ from \{$Solved$\} that is semantically closest to the current given question $q$. This process can be expressed as Equation \ref{eq:graph_ini}. Here, $SE(\cdot)$ takes the question sentence as input and outputs a fixed-length vector. $\|\ SE(q) - SE(q_s) \ \|$ judges the $L^2$ distance between $SE(q)$ and $SE(q_s)$. Then, the program found for $q_{c}$ in previous searches becomes the initial program for the Program Graph. 

\begin{equation}
    q_{c} \ = \  \mathop{\arg\min}_{q_s \in \{Solved\}} \|\ SE(q) - SE(q_s) \ \| 
\label{eq:graph_ini}
\end{equation}

\section{Legality Check for Programs}\label{apx:legal_check}

As stated in Section 3.3, 
our rules for generating mutations on programs can ensure the legality of structure, but not necessarily the legality of semantics. Here, the illegality of semantics mainly comes from the type system of modules. Within NMN, the inputs and outputs passed between modules are restricted with types such as feature map, number, object, or set of objects. The calculation of NMN fails if the intermediate data fed to a module does not match the data type that module requires. 

Generally, there are two solutions to this problem. One is to add the illegal programs to the Program Graph anyway yet mark these programs as non-executable and skip the step of trying these programs to get the accuracies. However, excessive illegal programs within the Program Graph waste plenty of searching steps on them so that the efficiency of search drops obviously. 

The other solution is to simply refuse to add these illegal programs to the Program Graph. However, in this way the Program Graph is possible to become disconnected. Therefore, some sub-graphs may never be reached from others.

In consideration of this, we applied a compromise between these two solutions. We used a hyperparameter named $Tolerance$ to restrict the maximum count of data type mismatches that can be tolerated. The programs of which the count of data type mismatches is not greater than $Tolerance$ will still be added to the Program Graph although they cannot be executed to obtain the accuracy. This setting can balance the efficiency and coverage of the search.

\section{Modules Used in FigureQA Dataset}\label{apx:modules}

The modules used in the experiment on the FigureQA dataset are shown in Table \ref{tab:modules}. 
Here, the column of ``Shape" indicates the number and type of inputs and output. 
The column of ``Architecture" indicates whether the module is pre-defined with rule-based calculation, or is a trainable neural network. 

\begin{table}[h]
\caption{Modules used in the experiment on FigureQA dataset}
\centering
\begin{tabular}{p{75pt}<{\centering} p{120pt}<{\centering} p{75pt}<{\centering} p{50pt}<{\centering}}
\hline
Name & Shape & Architecture & Number \\
\hline
Find Element & (None) $\rightarrow$ Element & pre-defined & 2 \\
Look Up & (Element) $\rightarrow$ Element & pre-defined & 1 \\
Look Down & (Element) $\rightarrow$ Element & pre-defined & 1 \\
Look Left & (Element) $\rightarrow$ Element & pre-defined & 1 \\
Look Right & (Element) $\rightarrow$ Element & pre-defined & 1 \\
Find Same & (Element) $\rightarrow$ Elements & pre-defined & 1 \\
Discriminator & \tabincell{c}{ (Element/Elements/None) * 2 \\ $\rightarrow$ Answer } & neural network & N \\
\hline
\end{tabular}
\label{tab:modules}
\end{table}

Specifically for the behavior of each module, 
``Find Element" finds an element that matches the given keyword from all the detected elements.
Here, keywords are the name of colors extracted from the questions. Because there are at most two keywords within a question, two of this module are required and each of them corresponds to one of the keywords.

``Look Up" finds the closest element that is in the area of from 45$^{\circ}$ top left to 45$^{\circ}$ top right of the given element. 

``Look Down", ``Look Left", and ``Look Right" behave similarly to ``Look Up". 

``Find Same" finds a set of elements with the same attributes as the given element. In this experiment, we specify this attribute to color. 

\begin{figure}[b]
\centering
\includegraphics[width=\textwidth]{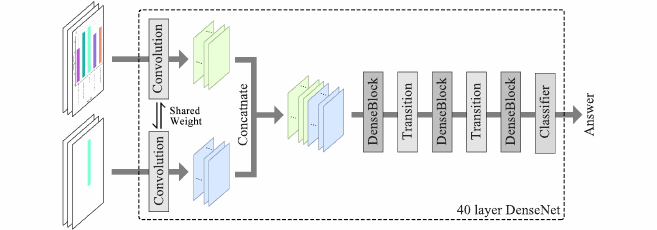}
\caption{Architecture of our Discriminator with a 40 layer DenseNet as backbone}
\label{fig:densenet}
\end{figure}

``Discriminator" has two inputs. For each input, it masks the original image with the bounding boxes of the given element or sets of elements. Then, the masked image is fed to a neural network to infer the answer. The input can also be empty. In this case, it would directly feed the original image to the neural network. 
To compare our method with existing work fairly, we use a 40-layer DenseNet similar to the one applied in PReFIL as the backbone of the Discriminator.
The architecture of Discriminator is shown in Fig.\ref{fig:densenet}. The number of filters in the first convolutional layer of DenseNet is 64. Considering that the two inputs of Discriminator are parallel and most of their features are similar, the first convolutional layer works on them independently with shared weight. All three following dense blocks have 12 layers. Their growth rate is set to 12. The number of final classes is 2 representing the answer ``Yes" or ``No" in FigureQA. 

For training, we used cross-entropy loss and SGD optimizer with learning rate decay. The batch size is set to 64. The learning rate is initialized to be 0.1 and drops to 0.01, 0.001, 0.0001, and 0.00001 on epoch 8, 12, 16, and 20, respectively. The maximum number of the epochs of training is 24, yet considering that the training of a 40-layer DenseNet on the entire 24 epochs is highly time-consuming, during the search only the training on the first 4 epochs are conducted and the validation accuracy is returned then. After the search on each question is completed, the Discriminator specified by the optimal program will be trained again on the entire 24 epochs.

\section{Results by Question Type and Figure Type in FigureQA Dataset}\label{apx:result_detail}

\begin{table}[htb]
\caption{Accuracy on Test Set 2 by different question types.}
\centering
\setlength\tabcolsep{1pt}
\begin{tabular}{l p{32pt}<{\centering} p{32pt}<{\centering} p{32pt}<{\centering} p{32pt}<{\centering}}
\hline
Question Template & RN & Human & PReFIL & Ours \\
\hline
Is X the minimum? & 76.78 & 97.06 & 97.20 & \textbf{98.44} \\
Is X the maximum? & 83.47 & 97.18 & 98.07 & \textbf{98.79} \\
Is X the low median? & 66.69 & 86.39 & 93.07 & \textbf{94.07} \\
Is X the high median? & 66.50 & 86.91 & 93.00 & \textbf{94.29} \\
Is X less than Y? & 80.49 & 96.15 & 98.20  & \textbf{99.43} \\
Is X greater than Y? & 81.00 & 96.15 & 98.07 & \textbf{99.45} \\
Does X have the minimum area under the curve? & 69.57 & 94.22 & 94.00 & \textbf{95.77} \\
Does X have the maximum area under the curve? & 78.45 & 95.36 & 96.91 & \textbf{97.65} \\
Is X the smoothest? & 58.57 & 78.02 & 71.87 & \textbf{80.90} \\ 
Is X the roughest? & 56.28 & 79.52 & 74.67 & \textbf{85.19} \\
Does X have the lowest value? & 69.65 & 90.33 & 92.17 & \textbf{95.42} \\
Does X have the highest value? & 76.23 & 93.11 & 94.83 & \textbf{96.68} \\
Is X less than Y? & 67.75 & 90.12 & 92.38 & \textbf{95.19} \\
Is X greater than Y? & 67.12 & 89.88 & 92.00 & \textbf{95.29} \\
Does X intersect Y? & 68.75 & 89.62 & 91.25 & \textbf{95.22} \\
\hline
Overall & 72.18 & 91.21 & 92.79 & \textbf{95.28} \\
\hline
\end{tabular}
\label{tab:result_qtype}
\end{table}

\begin{table}[htb]
\caption{Accuracy on Test Set 2 by different figure types.}
\centering
\begin{tabular}{c p{36pt}<{\centering} p{36pt}<{\centering} p{36pt}<{\centering} p{36pt}<{\centering}}
\hline
Figure Type & RN & Human & PReFIL & Ours \\
\hline
Vertical Bar & 77.13 & 95.90 & 98.25 & \textbf{98.55} \\
Horizontal Bar & 77.02 & 96.03 & 97.98 & \textbf{99.32} \\
Pie & 73.26 & 88.26 & 92.84 & \textbf{94.31} \\
Line & 66.69 & 90.55 & 87.79 & \textbf{92.66} \\
Dot Line & 69.22 & 87.20 & 89.57 & \textbf{93.11} \\
\hline
Overall & 72.18 & 91.21 & 92.79 & \textbf{95.28} \\
\hline
\end{tabular}
\label{tab:result_ftype}
\end{table}

\end{subappendices}

\end{document}